\pdfoutput=1

\documentclass[11pt]{article}

\usepackage{coling}

\usepackage{times}
\usepackage{latexsym}

\usepackage[T1]{fontenc}

\usepackage[utf8]{inputenc}

\usepackage{microtype}
\usepackage{url}
\hypersetup{breaklinks=true}
\usepackage{inconsolata}

\usepackage{graphicx}
\usepackage{tipa}
\newcommand{\umelb}{\textrm{\normalfont \textipa{@}}}
\newcommand{\uzur}{\textrm{\normalfont \textipa{Z}}}

\title{Can a Neural Model Guide Fieldwork? A Case Study on Morphological Data Collection}

\author{Aso Mahmudi$^\umelb$ \quad Borja Herce$^\uzur$ \quad  Demian Inostroza Améstica$^\umelb$ \\
   \textbf{ Andreas Scherbakov$^\umelb$ \quad  Eduard Hovy$^\umelb$ \quad Ekaterina Vylomova$^\umelb$} \\ 
    $^\umelb$The University of Melbourne  \quad  $^\uzur$University of Zurich\\
    \texttt{amahmudi@student.unimelb.edu.au} \quad \texttt{vylomovae@unimelb.edu.au}
    }

\begin{document}
\maketitle
\begin{abstract}
Linguistic fieldwork is an important component in language documentation and the creation of comprehensive linguistic corpora. Despite its significance, the process is often lengthy, exhaustive, and time-consuming. This paper presents a novel model that guides a linguist during the fieldwork and accounts for the dynamics of linguist-speaker interactions. We introduce a novel framework that evaluates the efficiency of various sampling strategies for obtaining morphological data and assesses the effectiveness of state-of-the-art neural models in generalising morphological structures. Our experiments highlight two key strategies for improving the efficiency: (1) increasing the diversity of annotated data by uniform sampling among the cells of the paradigm tables, and (2) using model confidence as a guide to enhance positive interaction by providing reliable predictions during annotation.
\end{abstract}

\section{Introduction}

According to UNESCO, around 2,000 languages are currently classified as endangered and over half of the languages spoken today might disappear by the end of the century.\footnote{\url{https://www.un.org/development/desa/indigenouspeoples/indigenous-languages.html}} In 2022, the organisation has declared the start of the decade of indigenous languages, and many linguists increased their efforts in documentation and revitalisation. But language documentation is a drawn-out, iterative, and exhausting process. A linguist would normally visit a language community several times to interview speakers and collect the data. During each visit, she or he would focus on tasks such as elicitation of words and language rules by offering them questionnaires or asking them to tell stories.  Between visits, the linguist would focus on processing, revising the data, and forming working linguistic hypotheses that will be further revised during the next face-to-face sessions. 

The amount of time spent in interaction with speakers is an important limiting resource, as native speakers often get tired in lengthy sessions, leading to a decline in their attention and interest, and, as a result, in poorer data quality \citep{bowern_linguistic_2015}.
\looseness=-1
\begin{figure}
\begin{center}
\includegraphics[scale=0.16]{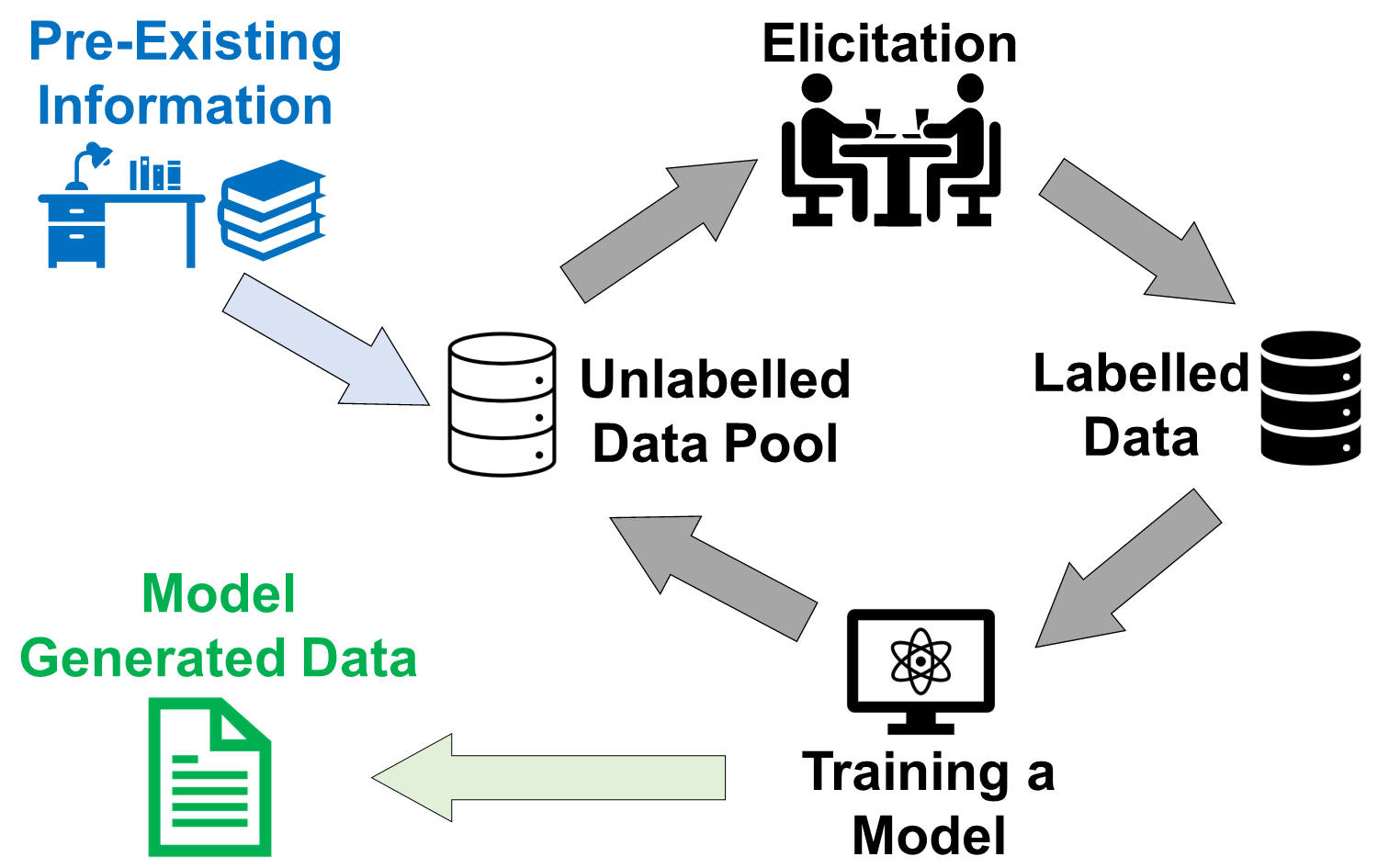} 
\caption{Illustration of the proposed word elicitation process model.}
\label{fig.Diagram}
\end{center}
\end{figure}

In this paper, we introduce \textbf{a neural system that guides the linguist, making the process of data collection more efficient}.\footnote{You can find all the code for this paper at \url{https://github.com/Aso-UniMelb/neural-fieldwork-guide}} The proposed model takes into account pre-collected data, identifies potential gaps in it, and informs the linguist of the (most informative) parts that should be collected in the next iteration. In contrast to existing approaches, we for the first time incorporate a measure that reflects an important ergonomic aspect of linguist-speaker interactions: we explicitly distinguish the following two cases of ``atomic'' linguist-to-speaker interactions:
(1) either a linguist makes a correct guess satisfying the speaker, or (2) seeks more information (e.g., upon producing ungrammatical utterances). The latter action tires the informant more than the former. Therefore, assuming that much greater cost associated to case (2) compared to case (1), we frame the planning of interaction sequences as an optimisation task. 

As a case study, we focus on morphological inflection data as it is characterised by high regularity and systematicity \citep{vylomova_compositional_2018} and neural models are particularly good at capturing regular patterns in data and have previously demonstrated high accuracy on morphological inflection shared tasks \citep{cotterell_conll-sigmorphon_2017, cotterell_conllsigmorphon_2018, mccarthy_sigmorphon_2019, vylomova_sigmorphon_2020, pimentel_sigmorphon_2021, kodner_sigmorphonunimorph_2022, goldman_sigmorphonunimorph_2023}.
As we aim to identify more data-efficient approaches, we also provide a comparative analysis of a variety of sampling strategies  (1) under a variety of data conditions as well as (2) in terms of their relevance and utility for the fieldwork pipeline.

For the first aspect, we include typologically diverse languages representing major morphological processes (fusion, agglutination), a variety of morphological complexities, and with ranging amounts of data available. For the second, we evaluate the models' ability to capture paradigm cell inter-predictability (discussed in Section~\ref{Experiments}). 

Our main contributions are:
 
1.  A novel approach to evaluate neural models that takes into account the nature of linguist-speaker interactions;

2. Evaluation of state-of-the-art models and sampling approaches for data-efficiency and ability to capture inter-cell predictability.

\section{Background} \label{sec:background}

\subsection{Motivation for the Word-and-Paradigm Model}
A key task in linguistic data collection involves the development and management of interlinear glossed texts, where morphological forms are broken down into units that carry meaning. While tools like ``FieldWorks Language Explorer (FLEx)''\footnote{ \href{https://software.sil.org/fieldworks/}{https://software.sil.org/fieldworks/} } offer some semi-automated assistance, interlinear glossing remains a highly time-intensive task for field linguists. The SIGMORPHON 2023 shared task on interlinear glossing \citep{ginn_findings_2023} highlighted efforts to automate this process and demonstrated that the availability of morphological segmentation plays a crucial role in achieving high accuracy. Still, morphological segmentation itself is a non-trivial task and a complicated problem in computational morphology \citep{batsuren_sigmorphon_2022}. 

An alternative method for morphological annotation is to adopt a model which does not necessitate segmentation. \citet{copot_word-and-paradigm_2022} also recommend a word-based approach to morphological annotation, especially for under-resourced and under-described languages. When working on a new language, a linguist collects and analyses wordforms, making generalisations about their relationships, and trying to identify morphological organisation, i.e., the structure and the size of the morphological paradigm (the number of paradigm cells). Having the paradigm structure, the linguist can then study the inter-predictability of the paradigm cells, trying to identify \textbf{principal parts},  the minimal subset of paradigm cells that provides all the necessary information to generate the other cells within the paradigm \citep{finkel_principal_2007}. In the well-known case of Latin, for example, all forms of the verb can be generated from just 4 forms \citep{finkel_what_2009}. Such knowledge allows for a more compact representation of linguistic rules and higher efficiency in data collection.

Many typical tasks in morphology such as paradigm discovery \citep{erdmann_paradigm_2020}, paradigm completion \citep{durrett_supervised_2013}, paradigm cell filling problem \citep{ackerman_parts_2009}, and morphological inflection \citep{kodner_sigmorphonunimorph_2022} are often approached using a word-based model. In theoretical linguistics, the Word-and-Paradigm model \citep{blevins2016word} offers a foundational framework for this word-based approach.

\subsection{Making the Data Collection Process More Efficient}
What is the best strategy to collect language data? As this process is time-consuming, it is essential to increase its efficiency. We explore active learning approaches in this paper. 
\textbf{Active Learning (AL)} has a well-established history in different NLP tasks \citep{zhang_survey_2022} and fits well with the language documentation process, where field linguists periodically consult with informants.
For instance, \citet{palmer_semi-automated_2009} used AL for real fieldwork experiments of a morpheme labelling task with two native speakers by examining three sequential, random, and uncertainty sampling strategies. \citet{muradoglu_eeny_2022} studied the simulated AL for a morphological inflection task on different languages with different sampling strategies. \citet{muradoglu_resisting_2024} found that the success of an inflection model on a test set largely depends on the entropy of the edit operations (required to transform a lemma into a target form) in the training data, and higher entropy which can be obtained by a uniform sampling across paradigm cells tends to improve the model's performance. \citet{erdmann_frugal_2020} proposed an approach to automate the paradigm cell filling problem task by manually providing a few forms. However, their method is impractical in real fieldwork settings because it requires the speaker (oracle) to frequently review the entire paradigm table.

\section{A Model of the Word Elicitation Process}
\textbf{Word Elicitation} is a technique used in linguistics to gather lexical and morphosyntactic data from native speakers with minimal contextual information. While corpora show what people {\em say}, elicitation uncovers what {\em can be said} \citep{meakins_understanding_2018}. To discover the morphological features, linguists usually change one feature at a time \citep{bowern_linguistic_2015}.
Elicitation cannot be sustained for an extended period in fieldwork, so it is recommended to limit it to around 20 hours spread across multiple sessions \citep{abbi_manual_2001}. In each session, the speaker is asked carefully designed short questions, and the linguist analyses the responses to generalise potential patterns.

This study focuses on modelling word elicitation during morphological data collection (as is illustrated in Figure~\ref{fig.Diagram}), with an emphasis on optimising process efficiency.

\subsection{Main Task and Initial Assumptions} 
\label{sec:assumptions}
The task involves filling in all plausible cells of the paradigm tables with correct inflected word forms. Cells that do not apply to specific lemmas are excluded from the process.

We assume the availability of pre-existing data, either gathered during early fieldwork stages or sourced from previous descriptive resources.

This data should include:

1) a basic word list (similar to the Swadesh list) consisting of verbs, nouns, adjectives, and other parts of speech provided in their dictionary forms (lemmas), and

2) a range of morphosyntactic features for each part of speech, which may be derived from prior studies or inferred from closely related languages, where applicable. We assume the knowledge of possible morphosyntactic feature combinations (tagsets such as ``\texttt{N;ACC;PL}'').

\subsection{Linguist--Speaker Interactions}
We now turn to the model of linguist-speaker interactions during the word elicitation process in morphological data collection.
We model a native speaker as an oracle system that has access to complete paradigms for all lemmas (labelled data pool). As an input, it receives (1) a lemma and (2) a target feature combination (tags corresponding to a paradigm cell).\footnote{In this work, we assume some linguistic expertise and knowledge of the features.} The linguist model is a neural system that can send requests to the speaker model. The requests might come at a certain cost as the process of word elicitation is exhausting, especially for native speakers \citep{bowern_linguistic_2015}. Whenever the linguist model retrieves a form or makes an incorrect prediction (in both cases the speaker model needs to return a valid form), it gets a penalty score of 1. In the case the linguist model checks a form and it is correct, the speaker is satisfied, and the linguist model does not get any penalty score. Hence, the linguist model has to optimise the retrieval process in order to minimise the penalty and increase the prediction accuracy. 

At some point, the linguist has to decide to stop the data collection process and return to their office. This means that they assume that the collected data is informative enough to accurately predict all the missing parts. Hence, at the final step, the linguist model predicts all the missing cells for each lemma. Whenever the prediction is incorrect, the model receives a penalty of 1 as well.

\begin{table*}[!ht]
\centering  \small
\begin{tabular}{lllllrrrr}
\textbf{Language} & \textbf{Code} & \textbf{Family} & \textbf{Typology} & \textbf{POS} & \textbf{Forms} &  \textbf{Lemmas} &	\textbf{APS} \\  \hline 
English         & \texttt{eng} & Germanic & analytic & V & 5,120 & 1280 & 4   \\ 
Latin           & \texttt{lat} & Romance & fusional  & V & 240,078 & 5,185 & 89  \\ 
Russian         & \texttt{rus} & Slavic & fusional   & N & 208,198 & 18,008& 16 \\ 
Central Kurdish & \texttt{ckb} & Iranic & fusional   & V & 21,375 & 375  & 57  \\ 
Turkish         & \texttt{tur} & Turkic & agglutinative  & V & 80,264 & 380  & 295\\ 
Mongolian       & \texttt{khk} & Mongolic & agglutinative   & N & 14,396 & 2057 & 8 \\ 
Central Pame    & \texttt{pbs} & Oto-Manguean & fusional  & V & 12,528 & 216  & 58  \\ 
Murrinh-patha   & \texttt{mwf} & Southern Daly & polysynthetic & V & 1,110 & 30  & 37  \\ 
\hline
\end{tabular}
\caption{Total number of wordforms, lemmas and average paradigm size (APS) for the selected part-of-speech (POS) across examined languages.} 
\label{tab:langs}
\end{table*}

\subsection{The Data Collection Model}

Once the initial data described in Section~\ref{sec:assumptions} is prepared, the linguist model generates for each lemma in the word list an unlabelled data pool. The pool consists of possible empty cells in the paradigm that correspond to plausible morphosyntactic feature combinations. 

As mentioned above, given the potentially large number of forms, it is impractical to ask the speaker model for all of them. Instead, a small subset of cells is selected over several rounds (cycles) of elicitation, and the linguist model is trained to generalise from that subset. The key here is to identify and target the most informative cells early on to gain a better understanding of the morphological structure.

Inspired by the 20-hour elicitation timeframe advised in fieldwork \citep{abbi_manual_2001}, and assuming 100 items are asked per hour, we limit our interaction to approximately 2,000 speaker (oracle) queries spread over five sessions, with 400 data wordforms retrieved in each cycle.

In the first cycle, the linguist model has no prior knowledge about the informativeness of each cell for facilitating generalisation and predicting other cells. At this stage, the model may either sample cells uniformly from the pool or start by gathering a few complete paradigms.

Note that in the latter option, the number of tables that can be collected from 400 queries will depend on their size in the corresponding language.

In some languages such as English, it might cover 100 paradigm tables, while in others, like Turkish, it might represent only two full paradigms (their average verbal paradigm size is greater than 200). 
Importantly, the availability of complete paradigms allows a linguist to infer cell inter-predictability and estimate the predictive power of each cell in paradigm tables and identify the principal parts.
In our experiments, we explore both strategies.

Once the initial processing is complete, the linguist needs to decide on the next cells to request from the speaker. Several strategies can be employed here: only checking the cells the linguist is most confident about (this reduces penalty but might be uninformative), exploring the most informative parts of the paradigm, or retrieving the cells with the highest uncertainty. We employ active learning \citep{ren_survey_2021} to optimise the sampling process. Each cycle here involves training a neural inflection model (a linguist model) to make generalisations about the data. While neural models typically require large amounts of data for training, they can generate predictions with varying levels of confidence at each training stage. We leverage this evolving capability to streamline interactions.

After several cycles of data collection, when we reach the approximate limit of 2,000 oracle queries, the trained neural model is used to predict the remaining pool data and its accuracy on these final predictions is evaluated.

\section{Experimental Setup}

\subsection{Datasets} \label{sec:data}
For this study, we selected 8 typologically diverse languages: English, Latin, Central Kurdish, Russian, Turkish, Khalkha Mongolian, Central Pame, and Murrinh-patha. The languages range in their morphological organisation, paradigm sizes, and levels of documentation. Table~\ref{tab:langs} provides a summary of the dataset specifications organised by language. 

The datasets are derived from UniMorph \citep{batsuren_unimorph_2022} and VeLePa \citep[Central Pame]{herce_velepa_2024}. The data samples are presented in the form of triplets consisting of a lemma (e.g., ``dog''), a target form (``dogs''), and morphosyntactic tags (``\texttt{N;PL}'').

\subsection{Experiments} \label{Experiments}

In our simulated data collection procedure, the oracle (speaker) is provided with access to the entire morphological dataset (labelled data pool). Additionally, for the remainder of the process, it also stores the forms that the linguist retrieved along with their predictions (if applicable). The linguist model has access to the data pool excluding the target form (i.e. unlabelled data). The linguist model, using its sampling strategy, selects a subset of lemma-target tag set combinations (a paradigm cell) from the pool and requests the corresponding target forms. When making a request to the oracle, the linguist model includes a predicted form if it has sufficient confidence in the prediction. If the prediction is correct, the oracle does not apply a penalty.

To evaluate sampling strategies and the interaction model, we design four experimental setups, which are described as follows.
In all experiments, the labelled data were collected over five cycles of AL, with 400 target forms gathered per cycle. The only exception is Murrinh-patha, where limited data availability required reducing the collection to 100 forms per cycle. Please note that whenever a neural model was trained, it was initialised from scratch and trained using all the data collected up to that point.

\textbf{Exp. 1:} 
In the first experiment, we model a baseline scenario when a linguist only asks a speaker to provide forms, without any particular strategy to select the most informative ones. Thus, here we uniformly sample a fixed number of cells from the pool in each of the five cycles. No suggestions were provided to the oracle throughout the experiment.

\textbf{Exp. 2:} 
In the second experiment, the linguist still does not have any particular sampling strategy but after the initial session, the linguist can make predictions with varying degrees of confidence based on observations from previous sessions and suggests the confident predictions to the speaker (hence reducing the chances of penalty). We modelled this case by using uniform sampling for each cycle and training a neural model on the collected data to provide confident predictions. The model predicted forms for all cells in the pool to determine an average confidence level. Subsequently, it retrieved the forms of randomly selected samples from the oracle and passed a prediction if its confidence surpasses the average confidence level.

\textbf{Exp. 3:} 
In the third experiment, a linguist collects some data, then studies it, and tries to fill in all the remaining cells in the whole data pool. Then they check with the speaker the forms they are most confident about and ask the speaker to provide forms they are puzzled about. This experiment follows a similar approach to the second, where a model was trained after the first cycle using random sampling. However, in the subsequent cycles, the sampling strategy was not random. The model generated predictions for the remaining pool data and ranked them based on confidence. Predictions with the highest confidence were queried from the oracle accompanied by a prediction, while the least confident predictions were obtained without one.
\begin{figure}
\begin{center}
\includegraphics[scale=0.45]{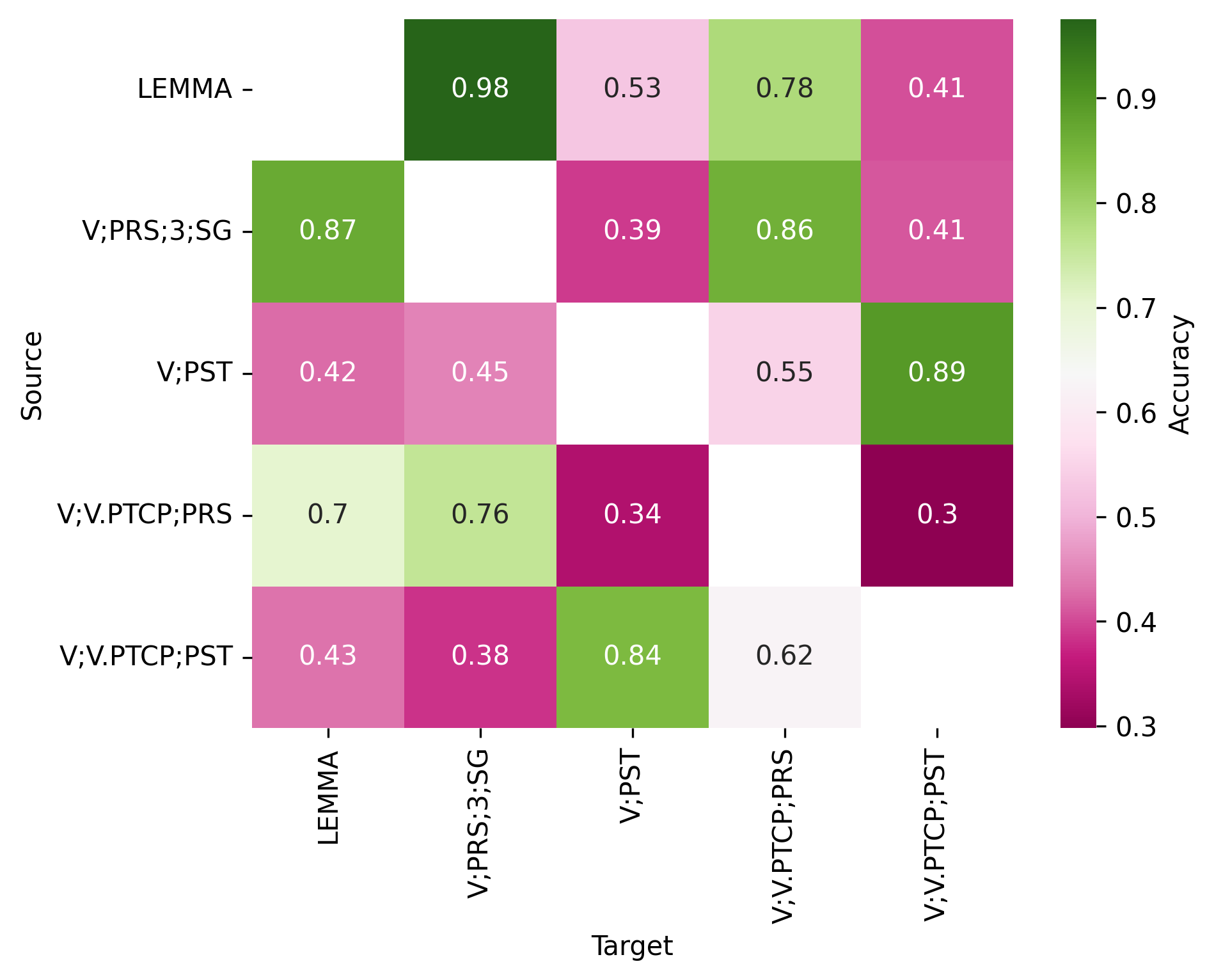} 
\caption{A heatmap showing the accuracy of predictions for English verbs.}
\label{fig.1}
\end{center}
\end{figure}

\textbf{Exp. 4:} 
The fourth experiment illustrates a scenario where the linguist first asks the speaker to complete full paradigms for a few lemmas. Then, the linguist assesses the inter-predictability of the cells to focus primarily on the cells with higher predictive power.
We describe this experiment in more detail as it introduces a novel method not previously explored. In the first cycle, the linguist model selects a small list of lemmas and asks the oracle for their complete paradigm table. The number of lemmas depends on the average size of the paradigm per language (assuming approximately 400 forms were queried). These data are used to identify the inter-predictability of cells in the paradigm tables.

We illustrate this process using English verbal paradigms due to its relatively small size. If we exclude the syncretic and non-morphologically realised forms, English paradigm tables would contain one lemma (the infinitive) and four inflected forms (present tense third person singular, simple past, past and present participle). Thus, we retrieve 400 English forms by requesting 100 paradigm tables, generate a dataset of all 2,000 possible re-inflection permutations (20 for each of the 100 verbs) and divide it into training, development, and test sets, with 45\%, 45\%, and 10\% of the data in each set, respectively. 
To explore the inter-predictability of cells, only once before the second cycle, we train a neural re-inflection model (details in Appendix~\ref{sec:appendix-model}) considering each cell as a source, aiming to predict from it the remaining forms in the corresponding paradigm table. We consider all possible source--target cell combinations, e.g. ``\texttt{went + V;PST + V;PRS;3;SG}'' was used as the input and ``\texttt{goes}'' as the output of the model to measure the predictability of ``\texttt{V;PST}'' with respect to ``\texttt{V;PRS;3;SG}'' for the lemma ``\texttt{go}''.  Figure~\ref{fig.1} shows a heatmap that indicates the model accuracy on the test set for different source and target tag combinations. The heatmap reveals that, in English, the lemma is generally a more informative source for predicting third-person singular present tense (``\texttt{V;PRS;3;SG}'') and present participle (``\texttt{V;PTCP;PRS}'') forms, compared to past tense (``\texttt{V;PST}'') or past participle (``\texttt{V;PTCP;PST}'') forms. Additionally, there is greater inter-predictability between simple past tense and past participle forms. The predictive power of an individual cell can be estimated from the average accuracy across the target cells. The system did not rely only on the most predictive cell. Instead, it employed these weights as fuzzy values in a weighted random sampling process. Based on these estimations, the system assigned weights for the remaining cells of the pool.

The sampling strategy for the following cycles of Exp.4 was similar to Exp.2, with the key difference being that in the second experiment, the sampling was uniform whereas in the fourth it was weighted random. The weights were determined by the estimated predictive power of each tagset. Like in Exp.2, a model was trained to predict the wordforms, and its predictions were passed to the oracle if the model had higher confidence in them.

\begin{figure}
\begin{center}
\includegraphics[scale=0.32]{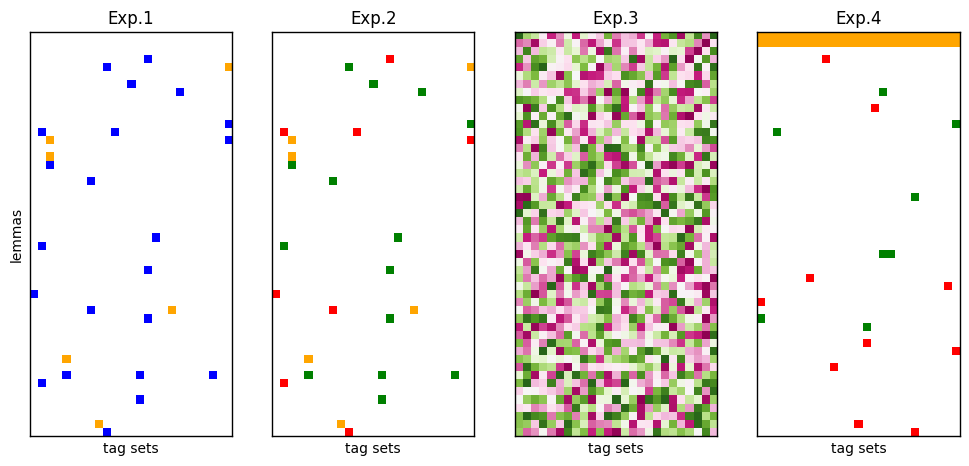} 
\caption{A simplified overview of sampling strategies used in the second cycle of the experiments. Blue cells represent samples retrieved without any predictions or confidence checks. Dark green cells denote confident ones retrieved with predictions, while dark red cells indicate low confidence cells with no predictions sent to the oracle. Orange cells indicate those that were selected in the first cycle and removed from the pool.}
\label{fig.experiments}
\end{center}
\end{figure}

To summarise the differences between the experiments, consider the second cycle illustrated in Figure~\ref{fig.experiments}. In Exp.1, cells were randomly selected for retrieval without any prediction. In Exp.2, the model passed predictions for confident cells, while no predictions for low confidence cells. In Exp.3, the most confident predictions were selected for retrieval with prediction, while the least confident ones were retrieved without prediction. Exp.4 followed a similar approach to Exp.2 but gave higher selection priority to more informative cells.

\section{Evaluation}
We evaluate the performance across the four experiments in terms of the accuracy of the final model and the efficiency of the process, using the following measures:

\paragraph{Accuracy on unseen data}
After the final cycle of the AL process, we calculate the accuracy of the inflection model trained on all retrieved samples in predicting the target form for the remaining samples in the pool (considering it as the test set).  

\paragraph{Normalised Efficiency Score}
We define a penalty score as an integer number by summing the number of times we call the oracle (excluding the times we propose a correct guess for the target form) and the number of incorrect predictions of the final model on the unseen test set.
Since the size of the datasets is not the same, we normalised the penalty by the total number of forms per language.
To better capture the efficiency of the elicitation process, we introduce a new metric—the complement of the normalised penalty—referred to as the Normalised Efficiency Score (NES).
This score is calculated as follows:
\begin{equation}
  \label{eq:NES}
  NES = 1 - \frac{P_{1} + P_{2} + P_{3} }{N}
\end{equation}
where \( P_{1} \) is the number of forms retrieved from the oracle without a suggestion, \( P_{2} \) is the number of forms retrieved with an incorrect suggestion, \( P_{3} \) is the number of incorrect predictions in the final test set, and  \( N \) is the total number of target forms in the dataset.

\section{Results and Discussion} \label{sec:results}
We conducted evaluation of the four experiments described in Section~\ref{Experiments}, across all the languages in our datasets.
For each iteration of active learning, the data labelled by the oracle was split into 90\% for training and 10\% for development.
This data was used to train an inflection model from scratch using a neural character-level transformer, following the hyper-parameters from \citet{wu_applying_2021}. At the end of each experiment, all remaining data in the pool was used as the test set and the final model predicted the corresponding target forms.

\subsection{Model Accuracy}
Table~\ref{tab:results-ACC} provides the target form prediction accuracy on the test set (the remaining samples in the pool) of examined languages.
Among the various sampling strategies tested in our experiments—uniform sampling, weighted random sampling based on estimated inter-predictability values, and sampling based on the model's confidence—uniform sampling yielded the highest prediction accuracy.
Our findings are consistent with previous studies \citep{muradoglu_eeny_2022,muradoglu_leveraging_2024}, confirming that random sampling across all paradigm cells is an effective strategy that cannot be outperformed easily when using smaller amounts of data, demonstrating its efficiency in the elicitation process. 

\setlength{\textfloatsep}{6pt}
\begin{table}
\centering \small
\begin{tabular}{l|llll}
lang & Exp.1 & Exp.2 & Exp.3 & Exp.4 \\ \hline
\texttt{tur} & \textbf{98.2} & 97.6 & 93.5 & 95.7 \\
\texttt{ckb} & 97.5 & \textbf{97.6} & 90.3 & 95.5 \\
\texttt{eng} & 89.2 & 89.0 & 89.0 & \textbf{90.9} \\
\texttt{khk} & 83.3 & \textbf{85.1} & 77.8 & 84.9 \\
\texttt{rus} & 84.2 & \textbf{85.8} & 71.1 & 84.3 \\
\texttt{lat} & \textbf{72.3} & 71.3 & 49.1 & 67.3 \\
\texttt{pbs} & 72.2 & \textbf{73.8} & 62.9 & 64.7 \\
\texttt{mwf} & \textbf{80.0} & 78.4 & 62.1 & 79.6 \\ \hline
Average & 84.6 & \textbf{84.8} & 74.5 & 82.9
\end{tabular}
\caption{Accuracy of the final model on remaining pool after the final cycle. Experiments 1 and 2 used identical sampling and their results are almost equal according to this evaluation metric.}
\label{tab:results-ACC}
\end{table}

\begin{figure*}
\centering
\includegraphics[scale=0.4]{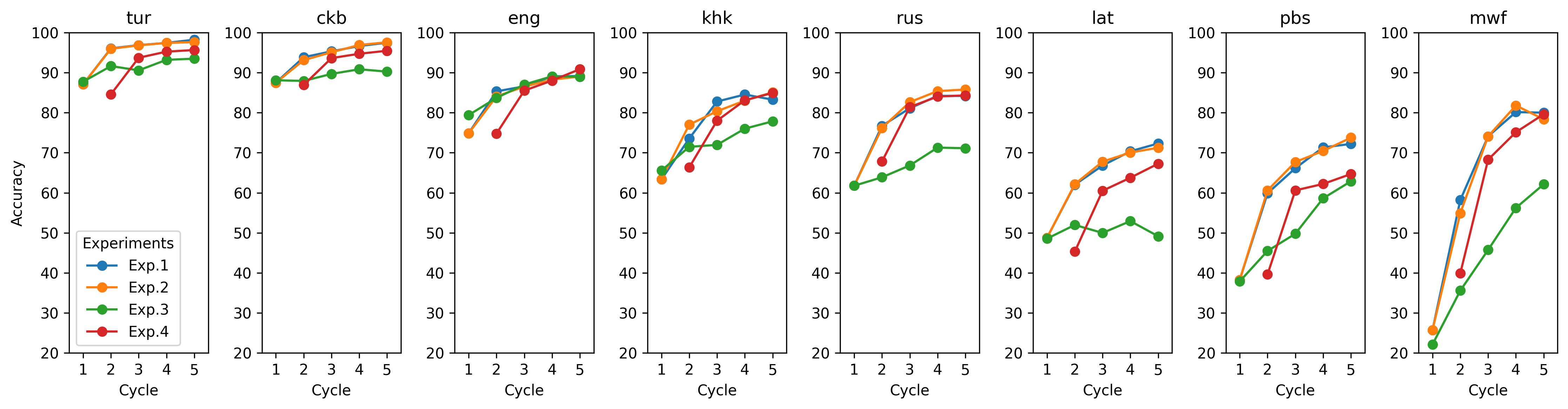} 
\caption{Accuracy on remaining pool data in each cycle of the active learning process for each language.}
\label{fig.cycles}
\end{figure*}

Next, we analyse the model's performance across active learning cycles. In all experiments, approximately 2,000 forms (500 for Murrinh-patha) were retrieved in total. Figure~\ref{fig.cycles} shows the accuracy of the inflection models on the remaining pool data in each cycle of the experiments. It demonstrates that accuracy improves with each cycle, initially increasing rapidly and then rising more slowly in the later cycles. However, Exp.3 shows limited accuracy gains for languages like Latin, Kurdish, and Russian. These languages have slots in their paradigms that either copy the lemma or exhibit regular consistent inflections. Confidence-based sampling tends to select these slots for providing suggestions, which restricts the diversity of the training data. This limitation is particularly evident in our Latin data, given its larger number of unique lemmas.

Due to the extremely low accuracy in the first cycle of Exp.4, we excluded them from Figure~\ref{fig.cycles}. This poor performance can be attributed to the limited lexical diversity of the training data, as most of it comes from just a few paradigm tables. However, in the third cycle, the accuracy in Exp.4, which used a weighted random sampling, improves significantly and approaches the performance of the uniform random sampling used in Exp.1 and Exp.2.

\subsection{Interaction Efficiency}
We now turn to an analysis of interaction efficiency. We observe that incorporating the confidence values of the inflection model for its predictions leads to sending more accurate predictions to the oracle, further enhancing the process's overall efficiency. Table~\ref{tab:results-NES} shows the normalised efficiency score for the experiments per language.

\begin{table}
\centering \small
\begin{tabular}{l|llll}
lang & Exp.1 & Exp.2 & Exp.3 & Exp.4 \\ \hline
\texttt{tur} & 95.8 & \textbf{96.3} & 92.5 & 94.1 \\
\texttt{ckb} & 88.4 & \textbf{92.4} & 87.0 & 90.3 \\
\texttt{eng} & 54.2 & 68.7 & \textbf{72.9} & 66.1 \\
\texttt{khk} & 71.7 & \textbf{78.2} & 72.0 & 76.4 \\
\texttt{rus} & 83.4 & \textbf{85.2} & 70.9 & 83.7 \\
\texttt{lat} & \textbf{71.7} & 70.9 & 49.2 & 66.9 \\
\texttt{pbs} & 60.7 & \textbf{66.0} & 58.2 & 57.3 \\
\texttt{mwf} & 44.0 & \textbf{54.4} & 48.3 & 49.6 \\ \hline
Average & 71.2 & \textbf{76.5} & 68.9 & 73.2
\end{tabular}
\caption{Normalised Efficiency Score of each experiment on different languages.}
\label{tab:results-NES}
\end{table}

\begin{figure*}
\centering
\includegraphics[scale=0.36]{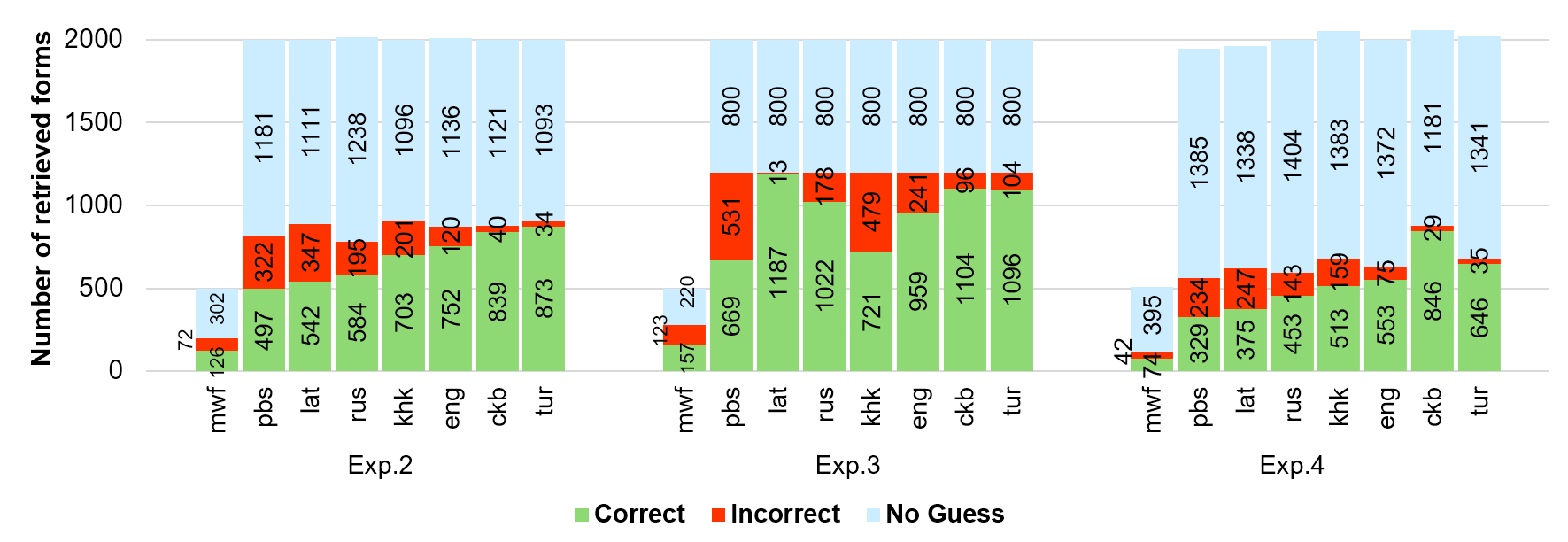} 
\caption{Submitted predictions along the requests to the oracle in each experiment. Exp.1 is omitted as all its requests were without a prediction.}
\label{fig.suggestions}
\end{figure*}

To better understand the interaction efficiency, we analyse the outcomes as follows: The linguist models (except in Exp.1), to minimise penalties, submitted their predictions with queries when sufficiently confident. Nonetheless, these predictions were not always accurate. Figure~\ref{fig.suggestions} illustrates the number of data samples retrieved from the oracle, segmented by the correctness of the submitted prediction. Exp.3 outperformed the others by employing a non-random sampling strategy based on the model's confidence. Overall, this demonstrates that, to some extent, we can rely on the model's confidence to enhance the efficiency of the interaction process.

To evaluate the impact of prioritising the completion of a few paradigm tables over the rest of the elicitation process, we designed Exp.4, where cell informativeness within paradigms was estimated and influenced the proportion of data retrieval. However, the results indicate that this approach does not significantly enhance the model's performance or efficiency, as successful generalisation in neural models largely depends on the lexical diversity and entropy of the training data.

\section{Conclusion}
In this paper, we evaluated neural models in their ability to guide fieldwork by accounting for the nature of linguist--speaker interactions in the process of language documentation. Focusing on morphological data collection,  we investigated various strategies for data sampling. Our results showed that uniform random sampling across paradigm cells results in more representative data and yields better generalisation in low-resource scenarios. Furthermore, we discovered that incorporating the model's confidence levels enhances interaction by guiding decisions on whether to send a prediction. This approach improves the process by offering predictions as suggestions during data annotation tasks.

\section{Future Work}
This study employed a simulated active learning approach for morphological data collection. To translate this into a real-world application, two user interfaces would be necessary: one for linguists to input existing data and another one for native speakers to provide the desired information.

Since native speakers may find complex tasks that require linguist knowledge tedious, we suggest that the linguist prepares a variety of simple sentences to change the user interface into fill-in-the-blank tasks. Naturally, designing these sentences is a challenging task that varies for each part of speech and requires some preliminary understanding of the language, which can be informed by the morphosyntactic features collected earlier. During the system's elicitation process, the speaker can fill in or correct the relevant part of the paradigm by considering the context and the lemma. For instance, to elicit the past tense of the verb `sleep' in English, the prompt could be ``\texttt{I [sleep] yesterday.}" 
This approach resembles the SIGMORPHON 2018 shared task 2 \citep{cotterell_conllsigmorphon_2018}.

In addition, to speed up the speaker data entry in the first cycle, the linguist can write some general rules as regular expressions to generate suggestions for each cell. Instead of typing from scratch, the speaker can accept the suggestion or make minor corrections where necessary.

If a required cell is not available for a word, the speaker should let the linguist know through the interface. The cell should be removed from the data pool and should be reviewed by the linguist later. For instance, if a noun is incorrectly labelled as a verb and the system requests its past form, its part of speech should be corrected.

Future studies could explore using inflection classes in evaluation or sampling strategies, though significant challenges remain. Defining the exact number of classes in each language requires considerable granularity, such as determining how many of them would be necessary to accurately predict irregular English verb forms —-- a matter on which linguists and educators may disagree. Additionally, resource limitations, especially in low-resource languages lacking comprehensive dictionaries or grammatical descriptions, hinder the identification of inflection classes for all lemmas.

\section*{Limitations}
We evaluated our method in a simulated manner across a variety of languages with different amounts of available data. We are assuming that our existing data (a wordlist, parts of speech, and morphological tags) are accurate and do not require any modifications during data collection. Additionally, we are assuming that the speaker does not make any errors during data entry. In real-life fieldwork scenarios, any type of error can occur, and a linguist should address them by making corrections as early as possible.

\section*{Ethics Statement}
We do not foresee any potential risks and harmful use of our work. Our analyses are based on licensed data which are freely available for academic use.

\section*{Acknowledgements}
This work was supported by 2024ECRG140. We also express our highest gratitude to John Mansfield, Khuyagbaatar Batsuren, and reviewers for their constructive and valuable feedback.

\bibliography{cited}

\appendix

\section{Model details} \label{sec:appendix-model}
You can find all the code associated with this paper at \url{https://github.com/Aso-UniMelb/neural-fieldwork-guide}. The implementation and setup details of the neural architectures used in this study are provided below for clarity and reproducibility.

1) Re-inflection Models (used only in Exp.4):
These models are one-layer Bidirectional Long Short-Term Memory (BiLSTM) networks implemented using PyTorch. The key hyperparameters used for training are:
\begin{itemize}
    \item Batch size: 16
    \item Hidden dimension: 256
    \item Learning rate: 0.005
    \item Training duration: 20 epochs
\end{itemize}
The training process utilises a specific method for embedding morphosyntactic tags. Instead of embedding each tag individually, the tags for each data sample are embedded as a single unit. This method ensures compact representations. The source tag set, input word, and target tag set are then encoded into a dense vector representation.

2) Inflection Models (All Experiments):
A neural character-level transformer architecture was employed to train the inflection models used across all experiments. This architecture follows the hyperparameters detailed in \citet{wu_applying_2021}. Transformers are particularly suited for this task due to their ability to capture long-range dependencies and complex relationships in inflection data. The character-level approach ensures a fine-grained understanding of morphological patterns at the subword level.
\end{document}